\documentclass[letterpaper, 10 pt, conference]{ieeeconf}  % Comment this line out if you need a4paper

\IEEEoverridecommandlockouts                              % This command is only needed if 
\usepackage{graphicx}
\usepackage{amsfonts}
\usepackage{bm}
\usepackage{algorithm}
\usepackage{algpseudocode}
\usepackage{caption}
\usepackage{subcaption}
\usepackage{amssymb}

\title{\LARGE \bf
Portable Multi-Hypothesis Monte Carlo Localization for Mobile Robots
}

\author{
\authorblockN{Alberto Garc\'ia}
\authorblockA{\textit{Intelligent Robotics Lab} \\
\textit{Rey Juan Carlos University}\\
aa.garciag@alumnos.urjc.es}
\and
\authorblockN{Francisco Mart\'in}
\authorblockA{\textit{Intelligent Robotics Lab} \\
\textit{Rey Juan Carlos University}\\
francisco.rico@urjc.es}
\and 
\authorblockN{Jos\'e Miguel Guerrero}
\authorblockA{\textit{Intelligent Robotics Lab} \\
\textit{Rey Juan Carlos University}\\
josemiguel.guerrero@urjc.es}
\and 
\authorblockN{Francisco J. Rodr\'iguez}
\authorblockA{\textit{Robotics group} \\
\textit{Universidad de León}\\
jfjrodl@unileon.es}
\and 
\authorblockN{Vicente Matell\'an}
\authorblockA{\centerline{\textit{Robotics group}} \\
\centerline{\textit{Universidad de León}}\\
\centerline{vicente.matellan@unileon.es}}
}

\begin{document}

\maketitle
\thispagestyle{empty}
\pagestyle{empty}

%%%%%%%%%%%%%%%%%%%%%%%%%%%%%%%%%%%%%%%%%%%%%%%%%%%%%%%%%%%%%%%%%%%%%%%%%%%%%%%%
\begin{abstract}

Self-localization is a fundamental capability that mobile robot navigation systems integrate to move from one point to another using a map. Thus, any enhancement in localization accuracy is crucial to perform delicate dexterity tasks. This paper describes a new location that maintains several populations of particles using the Monte Carlo Localization (MCL) algorithm, always choosing the best one as the system's output. As novelties, our work includes a multi-scale match matching algorithm to create new MCL populations and a metric to determine the most reliable. It also contributes the state of the art implementations, enhancing recovery times from erroneous estimates or unknown initial positions. The proposed method is evaluated in ROS2 in a module fully integrated with Nav2 and compared with the current state-of-the-art Adaptive ACML solution, obtaining good accuracy/recovery times.

\end{abstract}

\section{INTRODUCTION}

The fundamental capability of a mobile robot is navigation. Navigation allows a robot to use a map to locate itself, plan routes from origin to destination, and carry them out. Even though navigation has been working for decades, the problem is still far from being solved. One of the main issues is reproducibility: on the one hand,  it is hard finding robust algorithms that can be applied in a general way to different robots and scenarios.
On the other hand, finding localization algorithms in public repositories that are correctly implemented and easily replicated is not easy. Thus, it is not surprising that many works have not survived the robot for which they were designed, the laboratory in which they were implemented, or the paper in which they were presented. 

To solve the first limitation, this research proposes a localization method that allows Adaptive Monte Carlo Localization (AMCL)\cite{pfaff2006robust}, the current state-of-the-art localization algorithm, to become a global approximation capable of recovering from a situation of total ignorance, common at startup, and maintaining other hypotheses, useful in case of loss or hijacking. 

The novel contributions of this approach are:
\begin{enumerate}
    \item \textbf{A metric} to determine if a robot estimation, coded as an AMCL, is wrong. Usually, the uncertainty derived from the covariance matrix is used to determine when the estimation has deteriorated. Using uncertainty is ineffective since this deterioration could be due to other causes than a wrong estimate. Also, estimates with low uncertainty could be completely wrong. Our metric directly measures the consistency between the sensory measurements obtained with those expected for each particle of each estimation. We will demonstrate how this metric allows comparing two estimations effectively.
    \item \textbf{An observation model} based on geometric transformations and projections. This work is limited to a 2D map but perfectly supports 3D maps and sensors.
    \item \textbf{A dynamics of the collection of AMCLs}. We will show how our approach can maintain different estimations, created in positions consistent with the observations and continuously evaluated to eliminate erroneous ones or merge them if they converge to similar positions. 
\end{enumerate}

The Open Source development model faces the second limitation. It has brought many benefits to industry in general and to science in particular. Analyzing and executing publicly available implementations favors one of the principles of the scientific method: the reproducibility of scientific studies. In Robotics, ROS~\cite{quigley2009ros}, and the current version, ROS2~\cite{doi:10.1126/scirobotics.abm6074}, have greatly facilitated the reuse of other scientists' implementations by becoming the \emph{de facto} standard in robot software development, supporting hundreds of robots of different configurations and manufacturers. Nav2~\cite{macenski2020marathon2} is their navigation framework, which aspires to be the world's most widely used navigation system. The implementation of the work presented in this paper has been integrated into Nav2 so that it can be reused by other robotics researchers, guaranteeing that our method is general and applicable to any robot using Nav2.

This paper is organized as follows: After presenting the most related works to ours in section \ref{sec:related}, in section \ref{sec:amcl}, we will show the mathematical foundations of the AMCL used, emphasizing contribution 2. Section \ref{sec:mhamcl} brings together contributions 1 and 3, and presents how different estimations are maintained. The experimental validation is presented in section \ref{sec:validation}.

\section{RELATED WORK}
\label{sec:related}

Over the past two decades, multiple hypothesis localization algorithms have been developed. Jensfelt and Kristensen \cite{jensfelt2001active} proposed a hybrid localization method using Kalman filtering. Unlike the presented algorithm, the quantity of information extracted depends on the feature type, and a topological graph was used.

Yun and Miura \cite{yun2007multi} proposed a multi-hypothesis Kalman filter to generate and track the Gaussian pose hypothesis in outdoor environments. They used this technique to compensate for the lack of GPS precision in urban areas. The starting robot pose should be indicated. The number of pose hypotheses for each visual feature varies from 13 to 20. Another study was held in this kind of environment \cite{li2017estimating} also using a particle filter to develop a map matching method in autonomous vehicles incorporating additional information from external sensors, leaving the GNSS (Global Navigation Satellite Systems) influence at its minimum expression only used in the filter initialization and in the process of eliminating particles far away from the GNSS. The number of particles used was significantly higher: 1000. They also measured the GNSS error.

Davey \cite{davey2007simultaneous} presented a constraint-based tracking filter that employs the same technique as the constraint-based search in an interpretation tree for hypothesis tracking that can be applied to simultaneous localization and map building (SLAM). They stated that vision and laser sensor feature extraction could be reliable rather than a spatial feature. They identified data association as the main cause of a lost situation. Recent works such as the one proposed in \cite{xiong2022multi} formulated a SLAM technique using a radio-based ranging sensor, proposing a multi-hypothesis method that combines the advantages of the filter and graph optimization implemented via a multi hypothesis message passing process.

F. Martin et al. \cite{martin2007localization} proposed a Markovian method based on a Fuzzy Markov grid (FMK) for legged robot in the RoboCup soccer benchmark combined with a population of Extended Kalman Filters (EKFs), considering each EKF an independent hypothesis. The authors compared their contribution with a pure FMK and an FMK with a single EKF. The method mentioned above performed better than the other algorithms in terms of distance, orientation errors, and CPU time. However, regarding robot kidnapping, the recovery time increased significantly, contrasting to our algorithm.

There are also novel adaptations to AMCL, as in Chung and Lin's work \cite{chung2021improved}.
The authors propose an approach where the robot executes twenty times the ACML algorithm to avoid determining the estimation result in one loop. Thus, they get multiple algorithm estimations and determine one AMCL estimation based on the covariance matrix. This work maintains the hypotheses along the time and avoids high punctual computation rates associated with the multiple executions, which could finally have adversary effects on robot autonomy (e.g., battery).

\section{ADAPTIVE MONTE CARLO ALGORITHM} 
\label{sec:amcl}

This section will describe our Adaptive Monte Carlo Algorithm, highlighting the differences between the original algorithm and our proposal's \textbf{contributions 1 and 2}. We will focus on a single population of particles (a single AMCL). In the next section, we will address the management of various populations of particulate filters, corresponding to \textbf{contribution 3}.

AMCL applied to localization estimates the probability distribution $bel(x_t)$ that represents the robot's position at time $t$ as a set $\mathcal{S}_t$ of hypotheses about the robot's pose $\mathcal{X}_t$ (Equation ~\ref{eq:mcl1}). 

\begin{equation}
\label{eq:mcl1}
\mathcal{X}_t = \{x^1_t, x^2_t, \cdots, x^I_t\}
\end{equation}

Each hypothesis $x^i_t$ (with $1\leq i \leq I$) is a concrete instantiation of the robot's state at time $t$. 

$X_t$ has a $\mathcal{W}_t$ associated. Each hypothesis $x^i_t$ has an associated weight $w^i_t \in \mathbb{R}$ that denotes its probability given all
the previous perceptions $z_{1:t}$ and actuation commands $u_{1:t}$ (Equation \ref{eq:mcl2}).

\begin{equation}
\label{eq:mcl2}
w^i_t \sim P(x^i_t | z_{1:t}, u_{1:t}) 
\end{equation}

$\mathcal{P}_t$ is the set of \emph{particles}, being each \emph{particle} $p^i_t \in \mathcal{P}_t$ a tuple ${<x^i_t, w^i_t, h^i_t>}$. In the next section, we will explain what $h^i_t$ - let us ignore by now. For simplicity, when referring to a particle $p^i_t$, we can write ${<x, w, h>}$.

The robot position correspond to a normal distribution $\mathcal{N}(\boldsymbol{\mu}, \boldsymbol{\Sigma})$
with the mean $\boldsymbol{\mu}$ and covariance matrix $\boldsymbol{\Sigma}$ of $\mathcal{X}_t$.

%\begin{figure}[b]
%  \centering
%  \includegraphics[width=0.8\linewidth]{figures/TFs.png}
%  \caption{Frames relations.}
%  \label{fig:TFs}
%\end{figure}

The AMCL algorithm is divided into three phases in which the particles are updated to incorporate $z_t$ and $u_t$: 
\begin{itemize}
\item \textbf{Prediction}: Updates the positions of the particles $\mathcal{X}_t$ with the detected displacement $u_t$.
\item \textbf{Correction}: Updates the weights of the particles $\mathcal{W}_t$ with the sensor readings $z_t$.
\item \textbf{Reseed}: Removes hypotheses $x^i_t$ with weights $w^i_t < threshold$, creating new ones near to hypotheses with 
high weights.
\end{itemize}

The original algorithm specifies that these three phases are executed all the time sequentially. In our algorithm, we execute each phase independently at different frequencies to keep the computing load on the robot low. As reference: Prediction at 100Hz, correction at 10Hz, and reseed at 0.3Hz. The position is updated to high frequency so that when navigation decisions are made, the position is as accurate as possible. The particle weights can be updated less frequently because this is the most computationally expensive phase. It is unnecessary to carry out the reseed phase at high frequency without losing efficacy. The Adaptive Monte-Carlo Localization (AMCL) adapts the number of particles to the uncertainty of $\mathcal{X}_t$, that is, how concentrated or dispersed the particles are. When the uncertainty is low, the number of particles is reduced, increasing if the uncertainty goes up. 

% \section{PROPOSED ALGORITHM} 

% In the rest of the section, we will describe the main contributions of our proposal. We will begin by making a section on geometric reasoning, which is necessary to understand the formulation and algorithms of the other sections. 

% \subsection{Geometric reasoning}
Our work extensively uses a geometric transform system called TF that maintains the relationship (translation and rotation) between frames (reference axes). A coordinate in one frame can be transformed into another if these frames are connected in the same TF tree. A tree of TFs connects frames such that a frame has only one parent frame and can have multiple child frames.

As Figure \ref{fig:obsmodel} shows, the frame that represents the robot will be \emph{base\_footprint} (\emph{bf} to short it). There is a connection from \emph{bf} to the frame where all the laser readings are, \emph{base\_scan}. The frame \emph{odom} represents the point where the robot starts. The \emph{odom$\to$bf} transform encodes the displacement the robot has made since its start. The frame \emph{map} is the parent of \emph{odom}. A localization algorithm calculates the $map \to bf$ relationship, but since a frame can only have one parent, what is established after subtracting $odom \to bf$, is $map \to odom$.

The TF system receives a relation (rotation and translation) from which we obtain a $6 \times 6$ transformation matrix $RT^t_{A \to B}$, between two different frames from different sources, some of them at very high speed. To shorten, we will use the alternative notation $A2B_t$. This system can be asked for the relationship $X2Y_{t'}$ at time $t'$, even if $X$ is not directly connected with $Y$, and even if there is no information in the exact time $t'$. The TF system interpolates when necessary. 

\subsection{Prediction}
\label{sec:prediction}

The goal of this phase is to update the position of each particle $x^i_t\in X_t$
with the detected displacement $u_t$. If the last prediction was done in $t0$, the displacement $u_t$ is the difference
of the relations $odom2bf$ from time $t0$ and current time $t$ (Equation \ref{eq:mcl3}).

\begin{equation}
\label{eq:mcl3}
    u_t = odom2bf_{t0}^{-1} * odom2bf_t
\end{equation}

Each particle is updated (Equation \ref{eq:mcl4}) using $u_t$ plus a random noise $e(u_t)$ that follows a normal distribution $\mathcal{N}(0, E_u)$, being $E_u$ a parameter known \emph{a priori} that represent the odometry accuracy.

\begin{equation}
\label{eq:mcl4}
    x^i_t = x^i_{t-1} * (u_t + e(u_t))
\end{equation}

% \begin{algorithm}[h!]
% \caption{Prediction step}\label{alg:prediction}
% \begin{algorithmic}[1]

% \Function{e}{$u_t$}
%     \State \Return $u_t$ * random($\mathcal{N}(0, E_u)$)
% \EndFunction
% 
% \Function{Predict}{$\mathcal{P}_{t}$, $u_t$}
%    \ForAll{$x^i_{t}$ in $\mathcal{P}_t.\mathcal{X}_t$}
%         \State $x^i_{t} \gets x^i_{t-1} * (u_t + e(u_t))$
%     \EndFor
% \EndFunction
% 
% \end{algorithmic}
% \end{algorithm}

\subsection{Correction}
\label{sec:correction}

\begin{figure}[tb]
  \centering
  \includegraphics[width=0.6\linewidth]{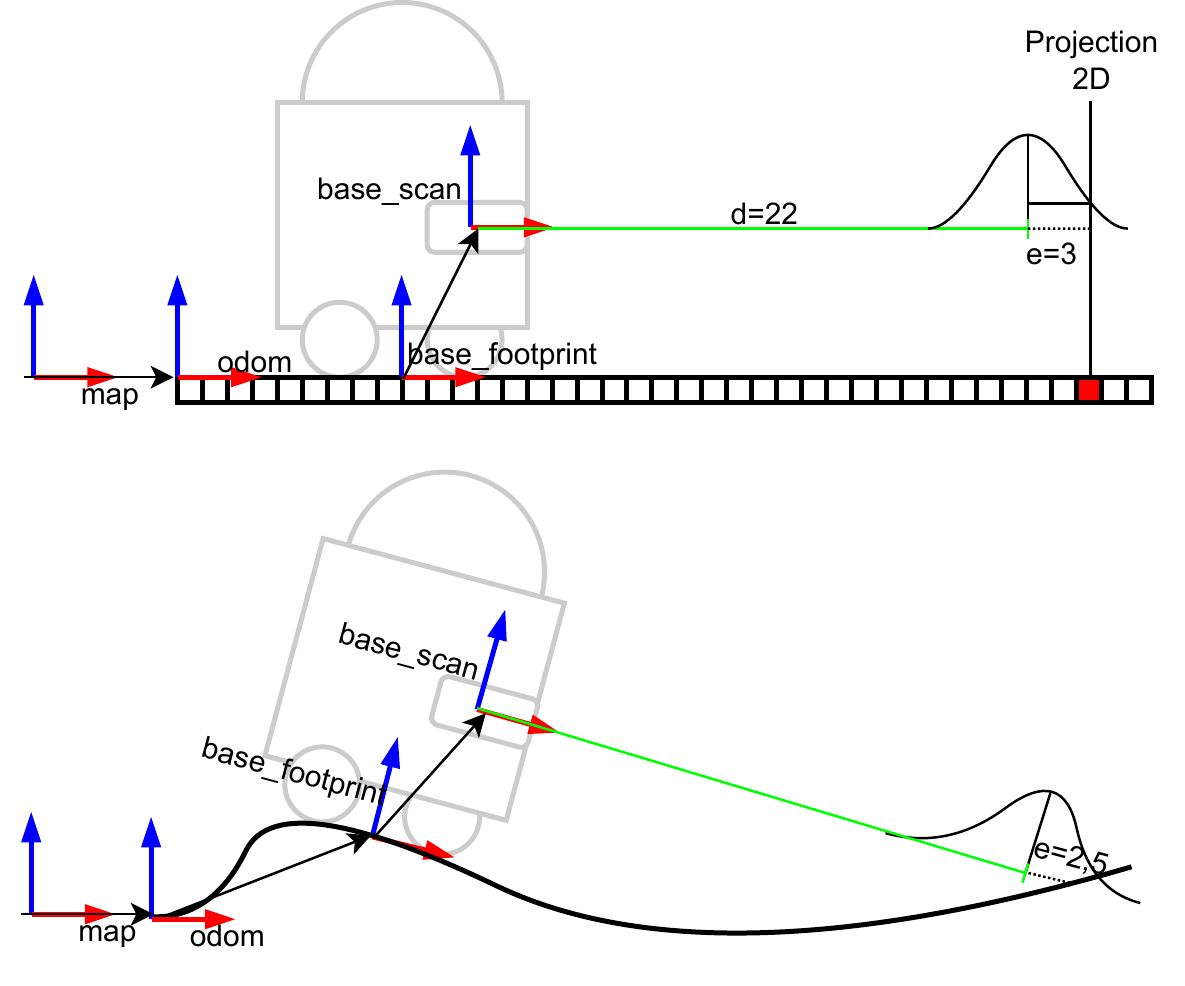}
  \caption{Observation model of a robot in a 2D cost map (up), and in a non-planar elevation/cost map (bottom).}
  \label{fig:obsmodel}
\end{figure}

In this phase, the observation model (\textbf{contribution 2} of our work), is applied to update $W_t$. As we discussed earlier, we heavily use the TF system, which geometrically models the relative positions between the sensor frame, the robot, the map, and the displacement from starting.
The robot is in a flat environment in this work, so the obstacle information is encoded in a 2D cost map. However, our approach would support other environments and encodings (Figure \ref{fig:obsmodel}). For this observation model, the calculations are performed in 3D,  projecting to 2D to obtain the cost of a cell. If we were to use an obstacle encoding using 3D voxels, or a non-planar environment, our approximation would also be valid.

In this work, we have used as observation $Z_t$ laser readings. Each of the reads $z^j_t \in Z_t$ has a $distance$ field and an $angle$ field, thus being in polar coordinates. In the correction step, each of the particles $p^i_{t} \in \mathcal{P}_t$ updates its probability $p^i_{t}.w$, comparing the sensory reading $Z_t$ with the one that should have been obtained if the robot was really in $\mathcal{P}_{t}.\mathcal{X}$. This comparison is made for each reading $z^j_t$ and for each particle $p^i_{t}$, using the Bayes theorem as shown by equation \ref{eq:mcl5} and \ref{eq:mcl6}.

\begin{equation}
\label{eq:mcl5}
    p^i_{t}.w = P(p^i_{t}.x | z^j_t) = \frac{P(z^j_t | p^i_{t}.x) * P(p^i_{t}.x)}{P(z^j_t)}
\end{equation}

\begin{equation}
\label{eq:mcl6}
    p^i_{t}.w = P(z^j_t | p^i_{t}.x) * p^i_{t}.w = \frac{1}{\sigma * \sqrt{2 * \pi}} e^{-\frac{1}{2}\frac{error}{\sigma}^2} * p^i_{t}.w
\end{equation}

The $\sigma$ parameter is a value that represents the sensor accuracy, and it is known \emph{a priori}. The \emph{error} value is the difference between the measured distance and the theoretical distance (Equation \ref{eq:mcl7}).

\begin{equation}
\label{eq:mcl7}
error = |z^j_t.dist - z'^j.dist|
\end{equation}

\begin{figure}[tb]
  \centering
  \includegraphics[width=0.45\linewidth]{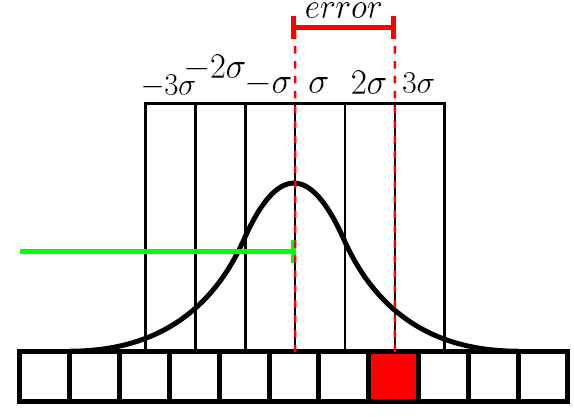}
  \caption{Calculation of the $error$ from the theoretical distance (green) and the obstacle in map.}
  \label{fig:obsmodel1}
\end{figure}

Obtaining the theoretical distance is very expensive computationally since the value of the costmap from the laser to the actual measured distance, and even beyond, should be consulted. Our sensor model follows a distribution $\mathcal{N}(dist, \sigma)$. Any obstacle at a distance greater than $3 \sigma$ from the theoretical one is practically zero (Figure \ref{fig:obsmodel1}). Our approach only queries the relevant cells to get this error.

Each particle in $\mathcal{P}_t$ has associated, in addition to a position $p^i_{t}.x$ and a weight $p^i_{t}.w$, a \emph{hits} field $p^i_{t}.h$ that indicates the likelihood of each particle with the last perception. The quality of $\mathcal{P}_t$ uses this field (Equation \ref{eq:mcl9}). In the validation section, we will show how this value is much more descriptive and reliable than using the covariance matrix.

\begin{equation}
\label{eq:mcl9}
Quality(\mathcal{P}_t) = \frac{\sum_{j=0}^n \frac{p^j_{t}.h}{|Z_t|}}{|\mathcal{P}_t|}
\end{equation}

\subsection{Reseed}
\label{sec:reseed}

In the reseed process, particles with less weight are replaced by particles close to those with more weight. There are two parameters: the percentage of winners and the percentage of losers. The process follows these steps:
\begin{enumerate}
    \item The population of particles $\mathcal{P}_t$ is ordered by their weight $p^i_{t}.w$, establishing which particles are \emph{winners} (particles with high weight) and which are \emph{losers} depending on the configured percentages. We can call the rest \emph{no-losers}.
    \item \emph{Loser} particles are removed.
    \item The same particles that have been eliminated are created, randomly selecting for each one a particle from the group of \emph{winners}, following a normal distribution $\mathcal{N}(0, winners/2)$.
\end{enumerate}

%\begin{figure}[tb]
%  \centering
%  \includegraphics[width=0.95\linewidth]{figures/reseed.pdf}
%  \caption{Reseed. Particles' color indicate high probability (green) to low probability (red).}
%  \label{fig:reseed}
%\end{figure}
We also take advantage of this step to fit the number of particles in the range $[particles\_min, particles\_max]$. If the distribution's covariance is high, we increase the number of particles. If it is low, we decrease it. With this, we reduce the computational load when the distribution has converged to a specific position.

\section{MULTI-HYPOTHESIS AMCL}
\label{sec:mhamcl}

We maintain a set $\mathfrak{P}_t$ containing ${\mathcal{P}^0_t, \mathcal{P}^1_t, ..., \mathcal{P}^N_t}$, being $N$ a parametrized limit.
The \textbf{contribution 3} of our work is the management of different $\mathcal{P}^k_t \in \mathfrak{P}_t$. The output of the location system is a position and a covariance, which corresponds to the $\mathcal{P}^k_t$ that is considered to fit the robot's position best. Skipping some details, the $\mathcal{P}^k_t$ whose $Quality(\mathcal{P}^k_t)$ is greater than the rest. We will describe this mechanism based on the following operations or phases:

\begin{itemize}
    \item \textbf{Start}: At the start of the robot operation, $P^0_{t0}$ is started at the robot's initial position, if known.
    \item \textbf{Creation}: Periodically, a cascade map matching algorithm is used to determine which map positions the latest sensory readings could be obtained. If a position with a high match is found, a new $P^k_t$ is started at this position.
    \item \textbf{Destruction}: If the $Quality(\mathcal{P}^k_t)$ is less than some threshold and $|\mathfrak{P}_t|> 0$, the $\mathcal{P}^k_t$ is considered to be wrong, it is removed.
    \item \textbf{Merge}:  Even if two $P^i_t$ and $P^j_t$, $i\neq j$, start at different positions, they could end up converging to the same position. In this case, they are mixed in a $P^k_t$ containing the particles with more weight.
\end{itemize}

%The process of managing the dynamic of $\mathfrak{P}_t$ is:

%\begin{enumerate}
%    \item At startup, $\mathfrak{P}_t$ is $\varnothing$, so we should start a $\mathcal{P}^0_t$ at the starting position.
%    \item If two $\mathcal{P}^i_t$ and $\mathcal{P}^j_t$ converge to the same position, all the particles of $\mathcal{P}^j_t$ are %stored in $\mathcal{P}^i_t$. Then the 50\% of the particles in $\mathcal{P}^i_t$ with lower quality are removed. Finally, %$\mathcal{P}^i_t$ is removed.
%    \item Low quality estimations are then destroyed.
%    \item Finally, the cascade map matcher returns the best candidate to start a new $\mathcal{P}^n_t$. If a candidate is far from %anyone in $\mathfrak{P}_t$, we create a new $\mathcal{P}^n_t$.
%\end{enumerate}

\subsection{Cascade Force Map Matching}

A map matching process is performed every few seconds to determine another position the robot might be in based on its sensory readings. We consider it \emph{brute} map matching because we calculate the likelihood between laser reading and the map for each cell map and each possible orientation.

\begin{figure}[tb]
  \centering
  \includegraphics[width=0.24\linewidth]{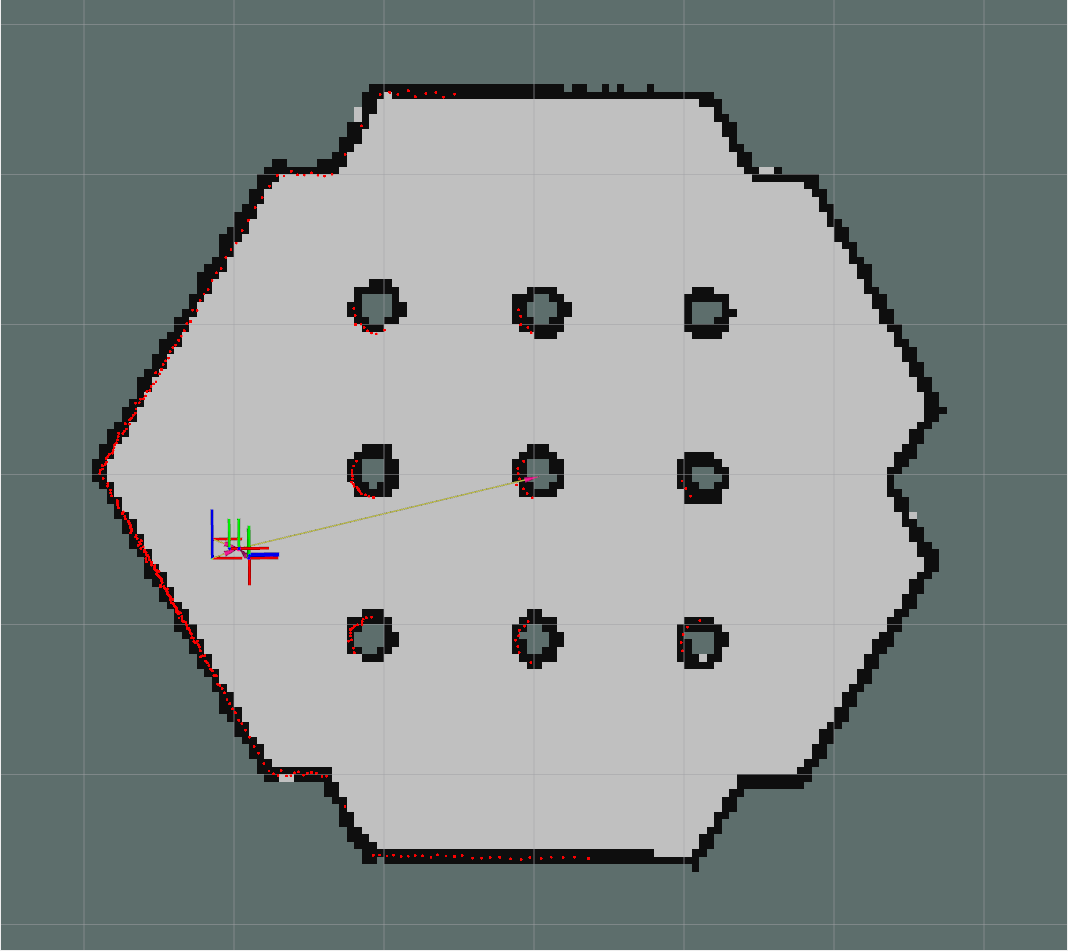}
  \includegraphics[width=0.24\linewidth]{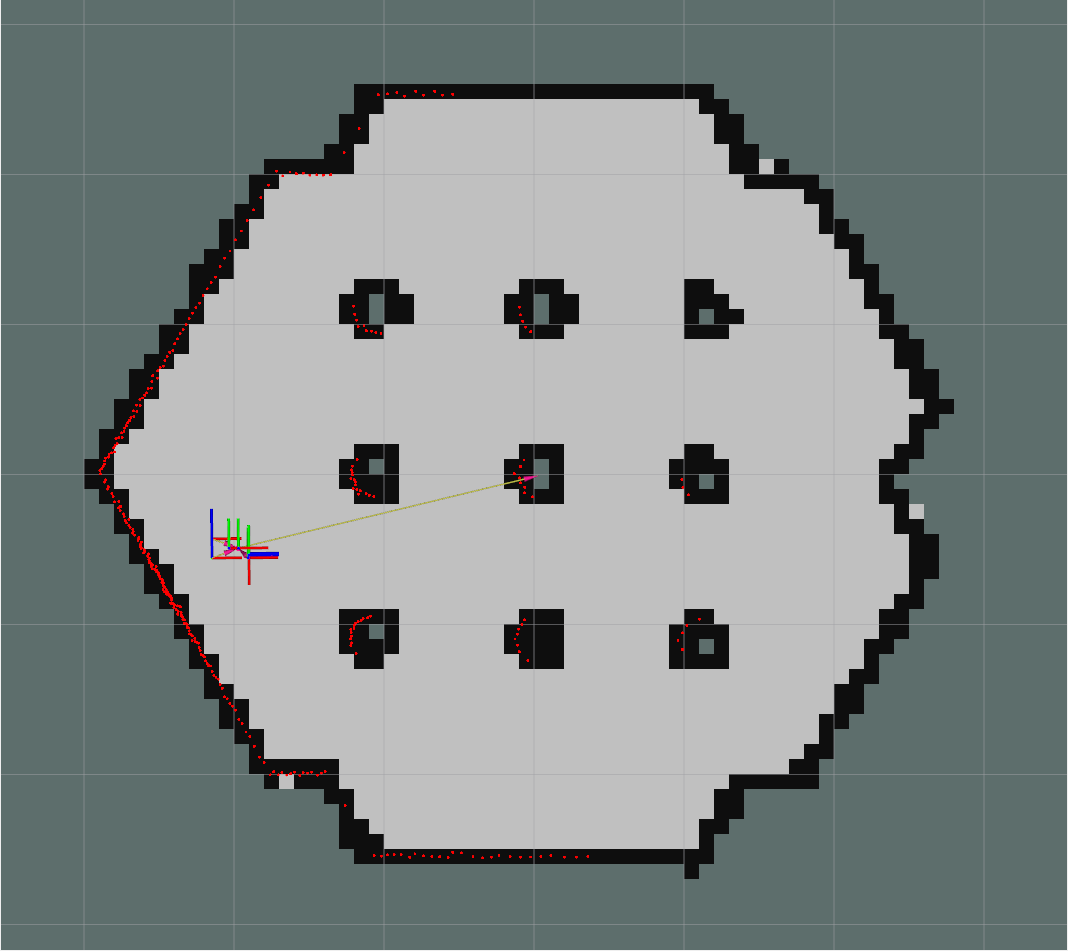}
  \includegraphics[width=0.24\linewidth]{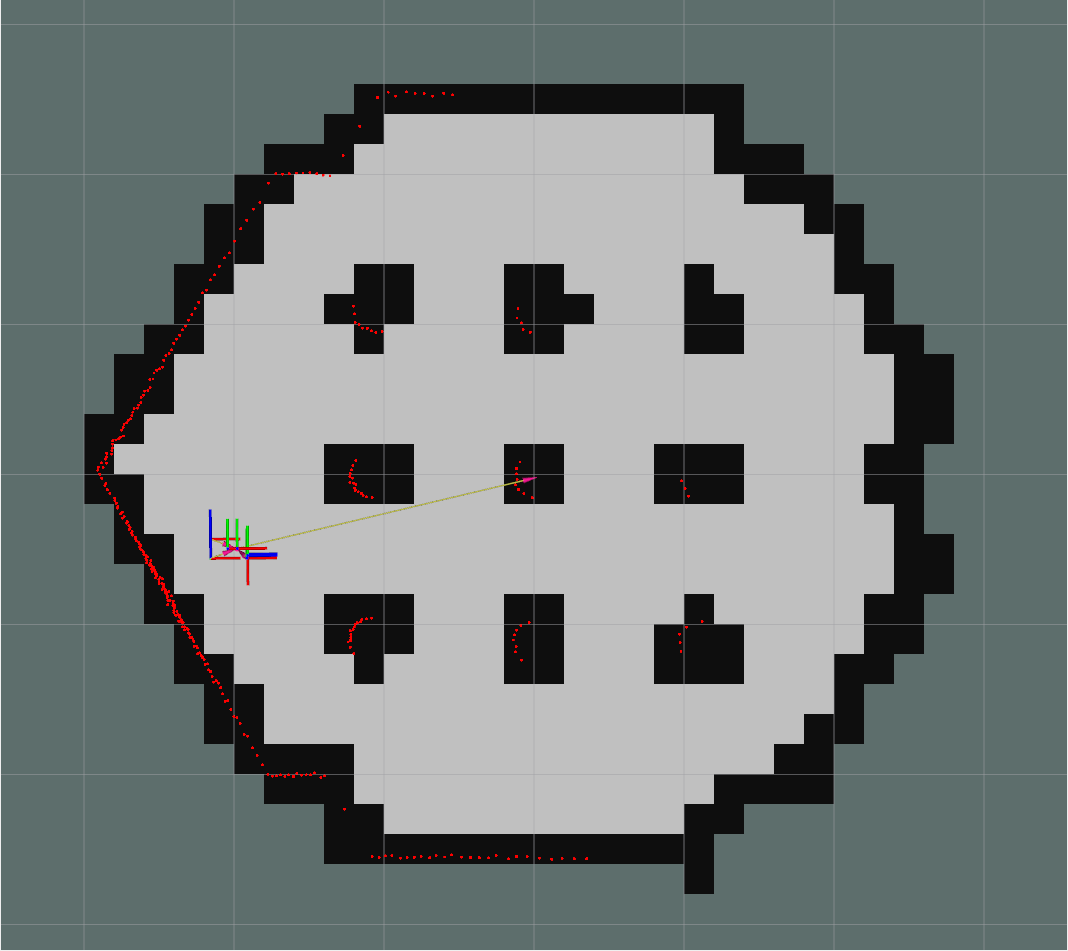}
  \includegraphics[width=0.24\linewidth]{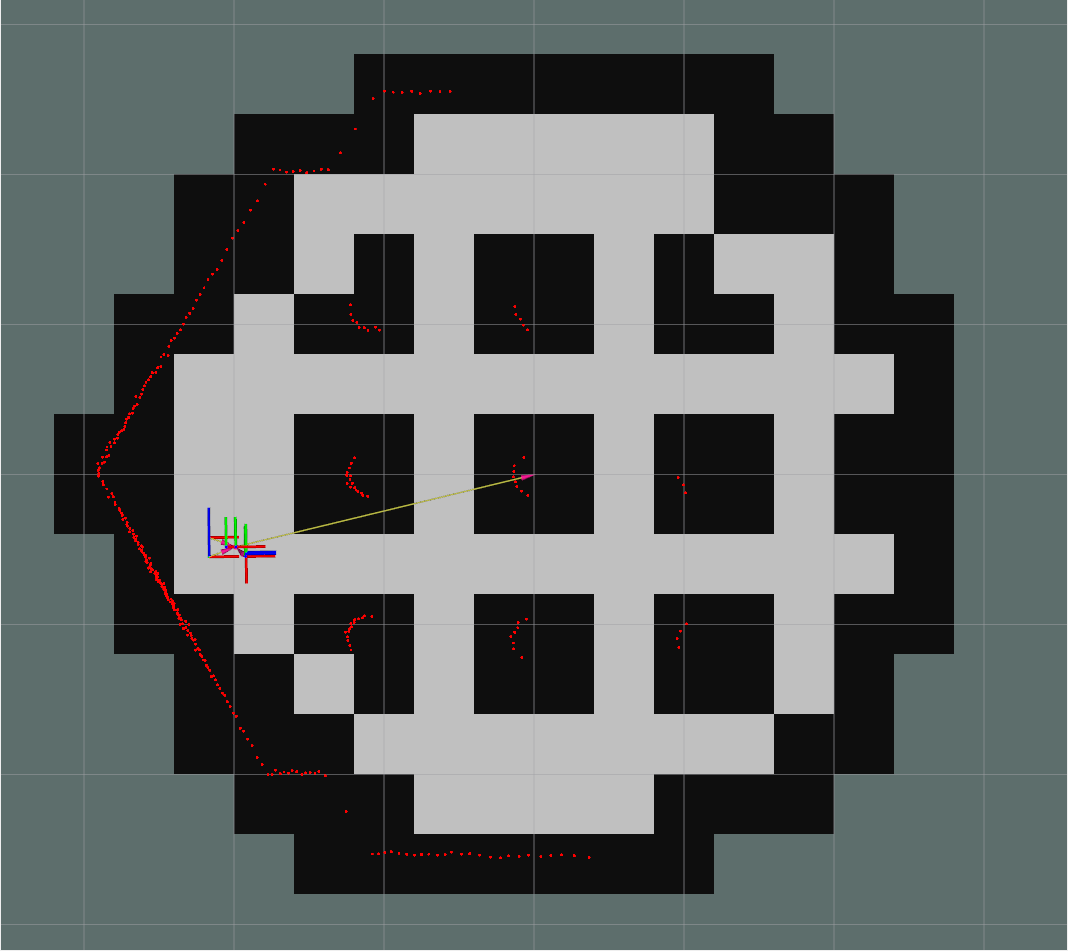}
  \caption{Map resolution levels used by the map matcher. Level 0 (top left) is the original resolution. The other figures correspond to incremental levels.}
  \label{fig:mh}
\end{figure}

Many calculations would consume all the robot's resources on large and/or high-resolution maps. To overcome this limitation, we perform this map math process by cascading different levels of resolution. At resolution level 0 is the original map. Each level duplicates the map resolution, as shown in Figure \ref{fig:mh}. Four levels are usually enough for most environments. Angular resolution is always $\frac{\pi}{8}$.

%\begin{algorithm}[h!]
%\caption{Manage Hypotheses}\label{alg:mh}
%\begin{algorithmic}[1]
%
%\Function{Manage\_Hypotheses}{$\mathfrak{P}_t$, $z_t$}
%    \If{$|\mathfrak{P}_t| = 0$}
%        \State Create($\mathcal{P}^0_{t0}$, init\_pos)
%    \EndIf
%    
%    \If{$|\mathfrak{P}_t| > 1$}
%        \For{$i=0$ to $|\mathfrak{P}_t|$}
%            \For{$i=0$ to $|\mathfrak{P}_t|$}
%                \If{$i \neq j$ and Near($\mathcal{P}^{i}_{t}$, $\mathcal{P}^{j}_{t}$))}
%                    \State $\mathcal{P}^{i}_{t0} \gets$ Merge($\mathcal{P}^{i}_{t}$, $\mathcal{P}^{j}_{t}$)
%                    \State Remove($\mathcal{P}^{j}_{t0}$)
%                \EndIf
%            \EndFor
%        \EndFor
%        \For{$i=0$ to $|\mathfrak{P}_t|$}
%            \If{$Quality(\mathcal{P}^{i}_{t}) < threshold$}
%                \State Destroy($\mathcal{P}^{i}_{t}$)
%%%            \EndIf
%        \EndFor
%   \EndIf
%    
%    \State $new\_pose \gets$ cascade\_map\_matcher($z_t$, $map$)
%    \If{$new\_pose \neq none$ and far($\mathcal{P}, \forall \mathcal{P} \in \mathfrak{P}_t$)}
%        \State $\mathcal{P}^n \gets create\_P(new\_pose)$     
%        \State $\mathfrak{P}_t \gets \mathfrak{P}_t \cup \mathcal{P}^n$
%    \EndIf
%    
%% Not finished
%\EndFunction
%
%\end{algorithmic}
%\end{algorithm}

We use the same metric shown in Equation \ref{eq:mcl9} to quantify if a position and orientation in the map is a promising candidate to start a new $\mathcal{P}^k_t$ on it. Starting from the higher level $N$, the ones with less resolution, we explore each map cell for the candidates. Each cell is evaluated with different orientations, with the angular resolution $\frac{\pi}{8}$. Once the candidates for level N are obtained, we repeat the same process in level $N-1$, but only in the cells corresponding to the candidates obtained in level N. We repeat the process until level 0, until getting a list with the resulting candidates sorted by quality.

\section{EXPERIMENTAL VALIDATION}
\label{sec:validation}

We have performed three experiments on a real robot to validate our contribution, taking the original AMCL algorithm~\cite{NIPS2001_c5b2cebf} in Nav2 as our baseline. This implementation has been ported from the ROS1 navigation framework to Nav2. It is a reliable, robust, and precise implementation, which thousands of robots have used for more than a decade in commercial and research applications, so we consider it impossible to find a more representative location algorithm to compare ourselves with.

The experiments carried out are:
\begin{itemize}
    \item \textbf{Experiment 1}. In this experiment, the robot starts from a known position and performs different itineraries in a scenario prepared in our laboratory (Figure \ref{fig:exp1scenario}). This experiment aims to measure our contribution's precision and study the dynamics of the different hypotheses when tracking the robot. 

\begin{figure}[b]
  \centering
  \includegraphics[height=0.33\linewidth]{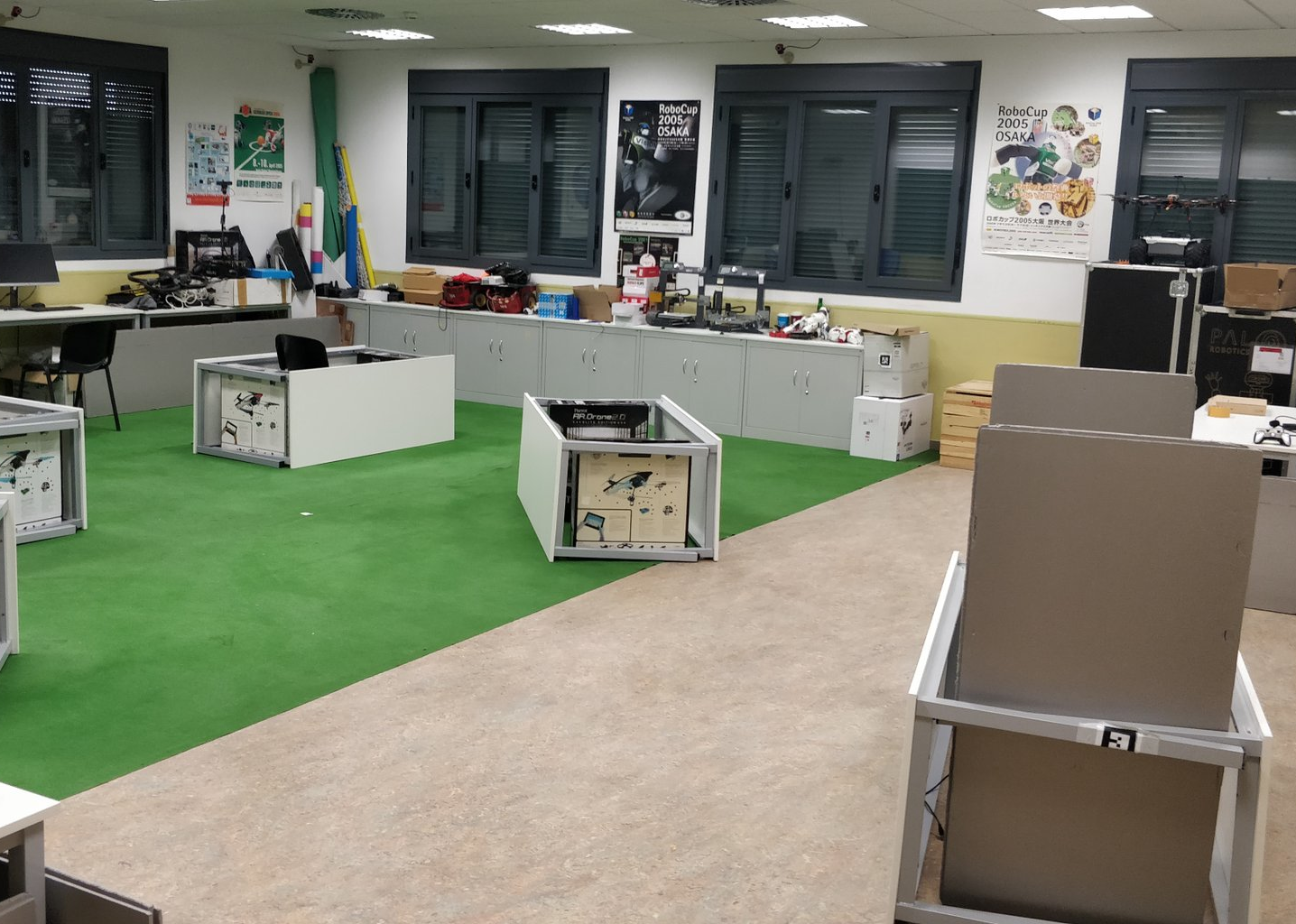}
  \includegraphics[height=0.33\linewidth]{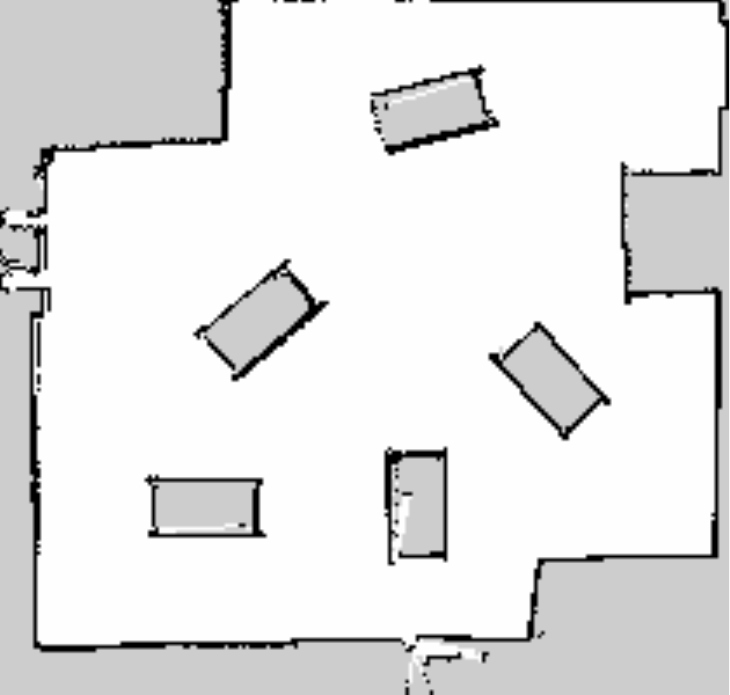}
  \includegraphics[height=0.33\linewidth]{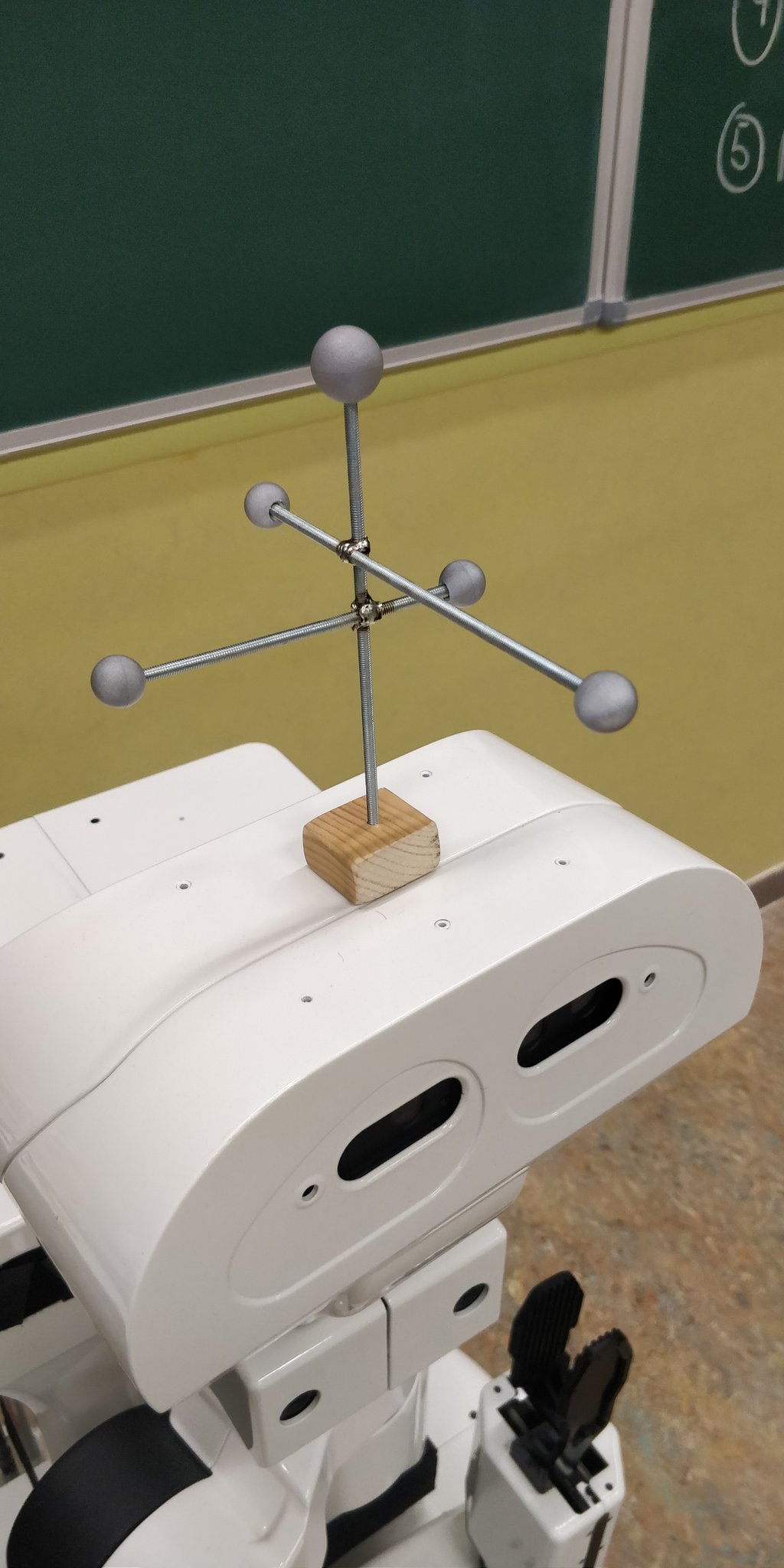}
\caption{Real scenario (left) and its map (center) for the first two experiments, and the motion capture object for tracking the robot (right).}
  \label{fig:exp1scenario}
\end{figure}

    \item \textbf{Experiment 2}. In the same scenario of the previous experiment, the robot starts from an unknown position, and we want to measure the ability to find its correct position on the map.
    \item \textbf{Experiment 3}. This experiment is carried out outside the laboratory in an uncontrolled environment where the robot performs long-distance and long-duration navigation. This experiment aims to show its robustness and effective integration with Nav2. This setup is the same as the one used in \cite{macenski2020marathon2}, using the same software\footnote{https://github.com/IntelligentRoboticsLabs/marathon\_ros2} to control and measure the experiment. The scenario is a long indoor corridor (100 meters long) with a round hub (top left).
    
%\begin{figure}[tb]
%  \centering
%  \includegraphics[width=0.98\linewidth]{figures/map_etsit.pdf}
%\caption{Scenario for the third experiment. The map represent a scenario with a round hub (top left) and long indoor corridor (100 meters long) in our campus.}
%  \label{fig:exp3scenario}
%\end{figure}
\end{itemize}

For the first two experiments, we recorded all the sensory and geometric information of the robot in some files called \emph{rosbags} (the rosbags of the experiments are available in our open repository\footnote{https://github.com/fmrico/mh\_amcl/tree/experiments/mh\_amcl/datasets} to be reproduced by the scientific community). This approach allows reproducing the experiment with different algorithms and parameters in the same conditions. These rosbags also include information on the ground truth of the robot, obtained using a motion capture system and a detection object (place in the robot's head in Figure \ref{fig:exp1scenario}), using our software MOCAP4ROS2\footnote{https://mocap4ros2-project.github.io/}, also available for further research.

%\begin{figure}[tb]
%  \centering
%  \includegraphics[height=0.39\linewidth]{figures/mocap.jpeg}
%  \includegraphics[height=0.39\linewidth]{figures/mocap2.jpeg}
%\caption{Motion capture system setup.}
%  \label{fig:mocap}
%\end{figure}

\subsection{Experiment 1: Tracking Robot}

In this experiment, the robot always starts from the origin of map coordinates. We have performed 5 iterations of the experiment with different itineraries. Each iteration lasts 2 minutes. The metrics we have used are:

\begin{itemize}
    \item \textbf{Accuracy}: Comparison between the real position of the robot and the one estimated by each algorithm. It is the most important metric when measuring a location algorithm.
    \item \textbf{Computing time}: Comparison of the CPU consumption of the algorithms to be measured.
    \item \textbf{Number of Hypotheses}:  How many hypotheses are held at a time in our algorithm.
    \item \textbf{Quality and uncertainty}: We will compare quality and uncertainty in our contribution to determine if quality may be a better metric when evaluating particulate filters.
\end{itemize}

\begin{figure}[h!]
  \centering
  \includegraphics[width=0.8\linewidth]{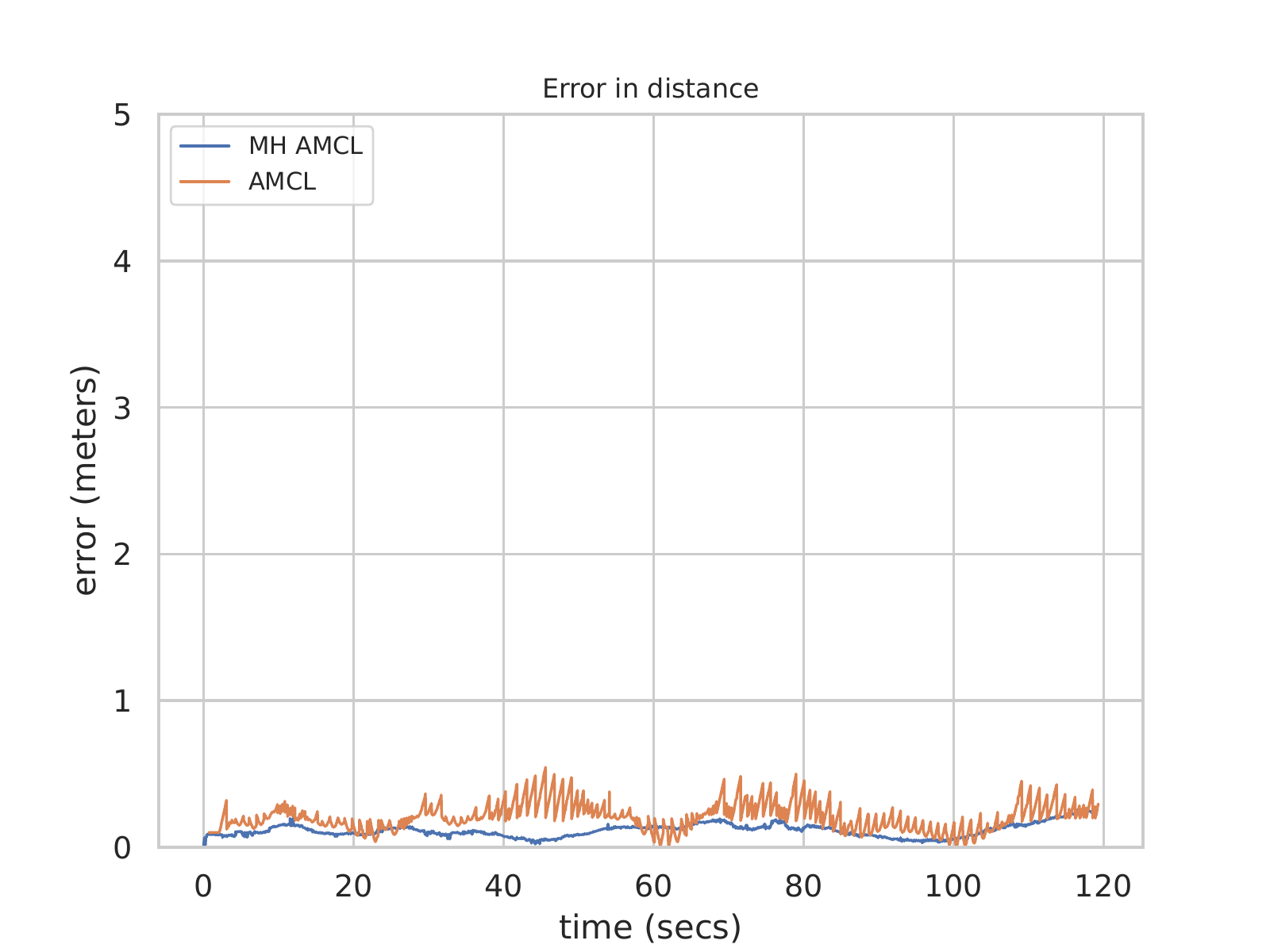}
\caption{Error in distance between MH-AMCL and AMCL}
  \label{fig:exp1error}
\end{figure}

Figure \ref{fig:exp1error} shows the accuracy of the algorithms, measuring the difference between the robot pose estimation and the real robot pose using the mocap system installed in our laboratory. We compare the accuracy between our contribution (MH-AMCL) and the original AMCL algorithm. Table \ref{tb:exp1} summarizes the results for all the experiment iterations, demonstrating that our proposal improves the estimation accuracy by reducing the error almost in half with respect to the baseline.

\begin{figure}[h!]
  \centering
  \includegraphics[width=0.8\linewidth]{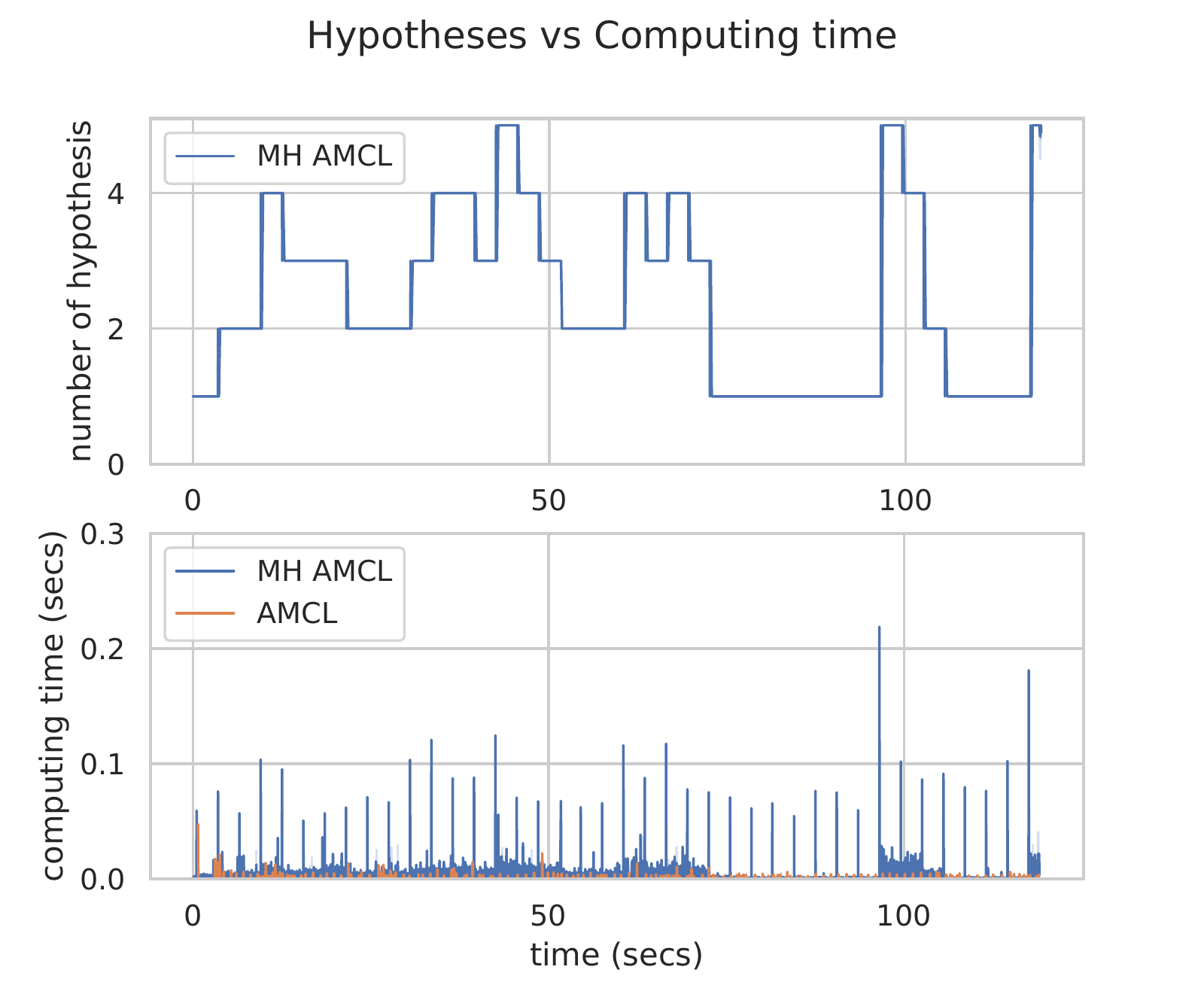}
\caption{Computing time of MH-AMCL and AMCL and number of hypotheses}
  \label{fig:exp1hypos}
\end{figure}

Figure \ref{fig:exp1hypos} shows a comparison of the computation cost of MH-AMCL and AMCL. The lower graph shows that our proposal increases the AMCL computation cost slightly. The peaks occur when the map matching process, triggered every few seconds, calculates new candidates. We can observe how the computation cost only increases lightly when the number of parallel estimations  ($|\mathfrak{P}_t|$) increases. As we can see in Table \ref{tb:exp1}, the increase of 2 milliseconds is worthless, and this algorithm is valid for real-time, increasing the accuracy almost double.

\begin{table}[]
\centering
\begin{tabular}{r|c|c|}
\cline{2-3}
\multicolumn{1}{l|}{}                & MH-AMCL & AMCL   \\ \hline
\multicolumn{1}{|r|}{Accuracy (m)}       & 0.109 $\pm$ 0.039  & 0.206 $\pm$ 0.098 \\ \hline
\multicolumn{1}{|r|}{Median (m)}         & 0.105   & 0.195  \\ \hline
\multicolumn{1}{|r|}{Computing time (s)} & 0.006   & 0.004  \\ \hline
\end{tabular}
\caption{Experiment 1. Comparison of the proposed Multi-Hypothesis AMCL, and original AMCL.}
\label{tb:exp1}
\end{table}

\subsection{Experiment 2: Recovery}

In this experiment, the robot starts from unknown positions in the environment. The principal metric is the \textbf{recovery time}, which is the time from the start of the experiment until the location error indicates that the robot has already been located.

We carried out this experiment 30 iterations from different positions. In 15 of them, the robot starts from a standstill and then moves. In the other 15, the robot is already moving when the experiment starts.

\begin{figure}[h!]
  \centering
  \includegraphics[width=0.75\linewidth]{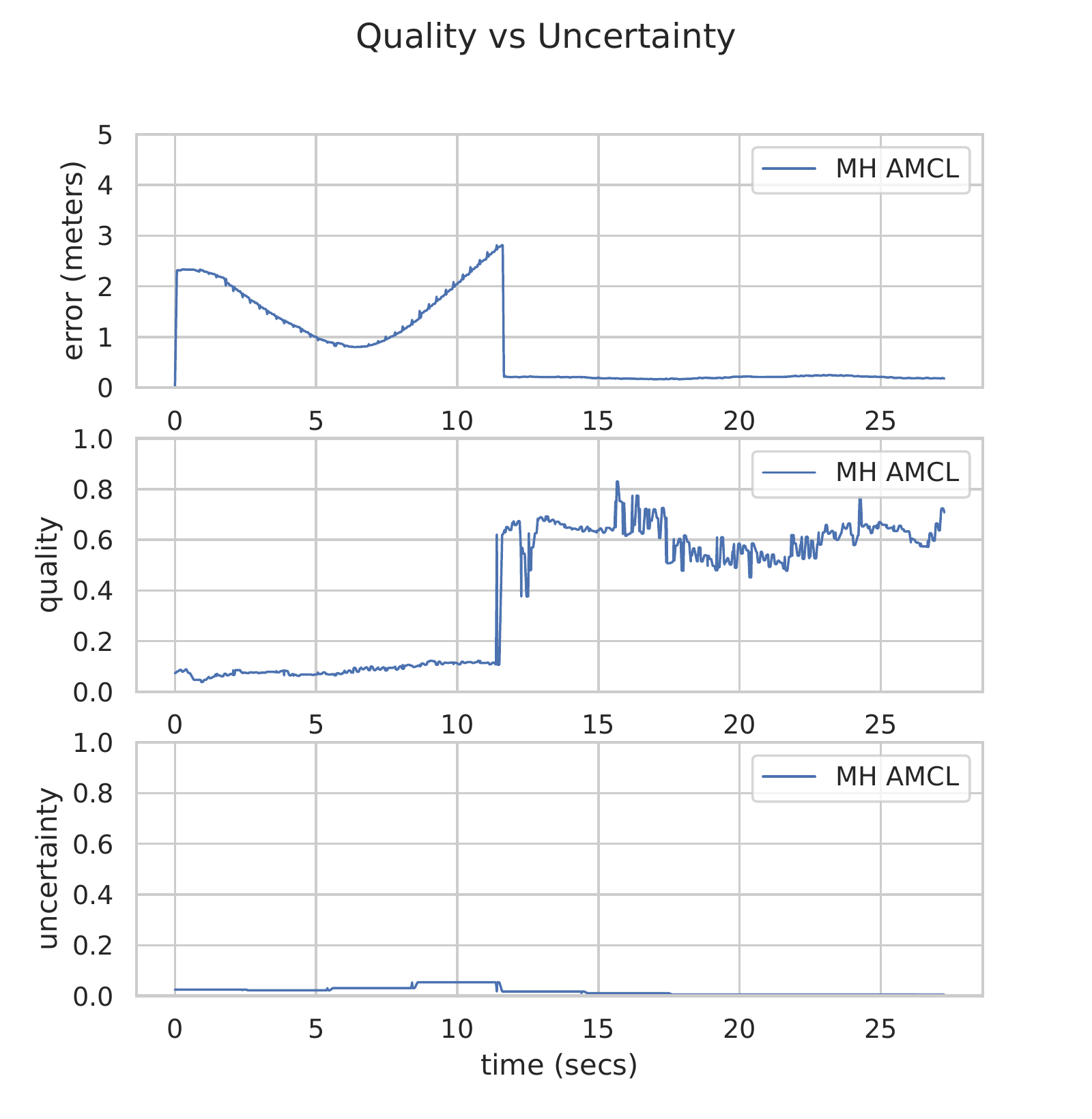}
\caption{Relationship between the accuracy, quality, and uncertainty}
  \label{fig:exp1quality}
\end{figure}

Figure \ref{fig:exp1quality} shows the relationship between the accuracy, quality, and uncertainty in one of the recovery tests. The first graph shows the error obtained between our contribution (MH-AMCL) and the mocap system. As observed, the error remains small and continuous until the end when the robot recovers from a wrong estimation. The second and third graphs show the quality and the uncertainty, respectively. Observe how the quality graph is low when the estimation is wrong and high when the estimation is correct. At the same time, the uncertainty is always low, making it useless to determine the recovery. From this, we can assume that uncertainty is not a good measure to detect if the robot is well-positioned. 

\begin{figure}[h!]
  \centering
  \includegraphics[width=0.75\linewidth]{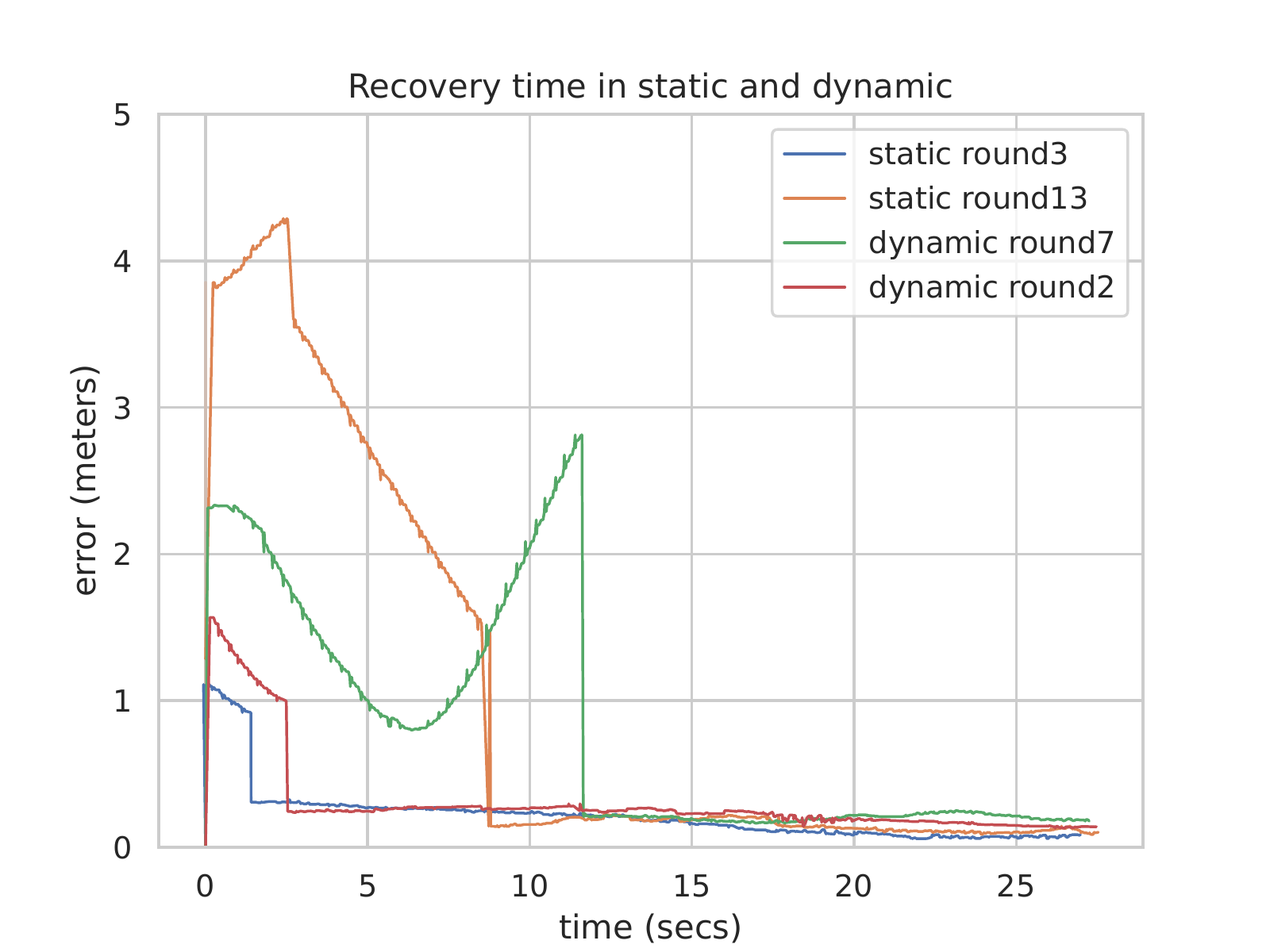}
\caption{Error during different recovery tests}
  \label{fig:exp1recovery}
\end{figure}

Finally, Figure \ref{fig:exp1recovery} shows the error in distance obtained in 4 representative iterations, two starting from a standstill (static) and two starts when the robot is moving (dynamic). As we can observe, when the error decreases abruptly and continues small until the end, the robot is well-located. The recovery times and success percentage of each MH-AMCL and the original AMCL are summarized in Table \ref{tb:exp2}. While the original algorithm takes an average of 13.714 seconds to locate with a success rate of 26.67\%, our contribution can locate the robot in all the iterations.

\begin{table}[h!]
\centering
\begin{tabular}{r|c|c|}
\cline{2-3}
\multicolumn{1}{c|}{}               & MH-AMCL & AMCL    \\ \hline
\multicolumn{1}{|r|}{Recovery time (s)} & 6.501   & 13.714  \\ \hline
\multicolumn{1}{|r|}{Success}       & 100\%    & 26,67\% \\ \hline
\end{tabular}
\caption{Experiment 2. Recovery time and success percentage.}
\label{tb:exp2}
\end{table}

\subsection{Experiment 3: Long-term navigation}

The objective of this experiment is only to show its integration with Nav2 and its robustness in its long-term operation in an environment with people around. We measured the \textbf{distance traveled}, the \textbf{time of the experiment}, the \textbf{recovery behavior} triggered during the experiment, \textbf{collisions}, and the \textbf{emergency stops} we had to make.

%\begin{figure}[h!]
%  \centering
%  \includegraphics[width=\linewidth]{figures/minimarathon.jpg}
%
%\caption{Robot executing our algorithm with Nav2 in a long-duration experiment.}
%  \label{fig:long}
%\end{figure}

The results are shown in Table \ref{tb:long}. The robot navigated for more than one hour, traveling 2.7 kilometers. There were 7 recoveries, all due to the presence of people in the surroundings. No collision, people damage, or emergency stops occurred during the experiment.

\begin{table}[h!]
\centering
\begin{tabular}{|l|c|}
\hline
\textbf{Distance  (meters)}      & 2734 meters \\ \hline
\textbf{Time (hrs)}              & 1.1         \\ \hline
\textbf{Recovery behaviors}      & 7           \\ \hline
\textbf{Num. of Collisions}      & 0           \\ \hline
\textbf{Num. of Emergency stops} & 0           \\ \hline
\end{tabular}
\caption{Experiment 3. Results for the long-term experiment.}\label{tb:long}
\end{table}

\section{CONCLUSIONS}

This paper has presented our location algorithm for mobile robots, whose main characteristic is that it can generate and manage new hypotheses about the robot's position in the environment. Each of these hypotheses is an independent Particle Filter, which competes with the others to be selected as the system output, based on a quality value that can determine which one is correct. Our work allows a robot to be unaware of its initial position or even to be able to locate itself again after a manual movement.

Our contribution has been integrated into the Nav2 navigation framework, guaranteeing its use out-of-the-box in any robot that supports ROS2 by other researchers or companies that require its features. It has been validated in a real robot in a controlled laboratory environment, using a motion capture system, and in an uncontrolled environment with people in a long-term test.

We are extending this work in several directions. Our approach is prepared to use 3D and elevation maps instead of 2D occupancy maps since the implementation heavily uses the ROS2 TF transform system and the particle position, which is modeled in 3D. Our future efforts also aim to improve the map matching algorithm further to support 3D maps and streamline computation. Finally, we want to explore encoding each hypothesis as an NDT-MCL instead of an AMCL.

%\section*{ACKNOWLEDGMENT}
%
%This work has been partially funded by Ministerio de Econom\'ia and Competitividad of the
%Kingdom of Spain under project PID2021-126592OB-C22 and by European Comission under grant XX-XXXX-XX.

\bibliographystyle{IEEEtran}  
\bibliography{references}  

\end{document}